\documentclass{article}
\usepackage{ijcai24}

\usepackage{times}
\usepackage{soul}
\usepackage{url}
\usepackage[hidelinks]{hyperref}
\usepackage[utf8]{inputenc}
\usepackage[small]{caption}
\usepackage{graphicx}
\usepackage{amsmath}
\usepackage{amsthm}
\usepackage{booktabs}
\usepackage{algorithm}
\usepackage[switch]{lineno}
\usepackage{makecell}

\usepackage{lipsum}
\usepackage{amsmath}
\usepackage{empheq}
\usepackage{amssymb}
\usepackage{algpseudocode}
\usepackage{amsthm}
\usepackage{bm}
\usepackage{appendix}

\newtheorem{thm}{Theorem}
\newtheorem{lemma}{Lemma}

\def\R{\mathbb{R}}

\newcommand{\w}{\bm{w}}

\newcommand{\x}{\bm{x}}

\newcommand{\y}{\bm{y}}
\newcommand{\W}{\bm{W}}

\newcommand{\I}{\mathbb{I}}
\newcommand{\f}{\bm{f}}

\title{Provable Acceleration of Nesterov's Accelerated Gradient Method over Heavy Ball Method in Training Over-Parameterized Neural Networks}
\author{
Xin Liu$^{1}$
\and
Wei Tao$^{2}$\thanks{Corresponding Authors.}\and
Wei Li$^1$\and
Dazhi Zhan$^1$\and
Jun Wang$^1$\and
Zhisong Pan$^{1\ast}$
\affiliations
$^1$Army Engineering University of PLA\\
$^2$Strategic Assessments and Consultation Institute, Academy of Military Science\\
}

\begin{document}

\maketitle

\begin{abstract}
Due to its simplicity and efficiency, the first-order gradient method has been extensively employed in training neural networks.
Although the optimization problem of the neural network is non-convex, recent research has proved that the first-order method is capable of attaining a global minimum during training over-parameterized neural networks, where the number of parameters is significantly larger than that of training instances.
Momentum methods, including the heavy ball~(HB) method and Nesterov's accelerated gradient~(NAG) method, are the workhorse of first-order gradient methods owning to their accelerated convergence.
In practice, NAG often exhibits superior performance than HB. 
However, current theoretical works fail to distinguish their convergence difference in training neural networks.
To fill this gap, we consider the training problem of the two-layer ReLU neural network under over-parameterization and random initialization. 
Leveraging high-resolution dynamical systems and neural tangent kernel~(NTK) theory, our result not only establishes tighter upper bounds of the convergence rate for both HB and NAG, but also provides the first theoretical guarantee for the acceleration of NAG over HB in training neural networks.
Finally, we validate our theoretical results on three benchmark datasets.
\end{abstract}

\section{Introduction}
Nowadays, the first-order gradient method is widely adopted as the standard choice for training neural networks~\cite{DBLP:journals/nn/Schmidhuber15}.
The gradient descent~(GD) method is the most famous first-order method, which only accesses the current gradient to update the model. 
To accelerate its convergence, the momentum method is proposed by utilizing the historical gradients.
The idea of incorporating momentum term into GD can be traced back to 1964, when Polyak introduced the heavy ball~(HB) method for solving the quadratic convex optimization problem~\cite{polyak1964some}.
Compared to GD, HB achieves a local accelerated convergence rate in a neighborhood around the global optimal.
Subsequently, Nesterov proposed another important type of momentum method known as Nesterov's accelerated gradient~(NAG)method~\cite{nesterov27method}.
In the convex setting, NAG achieves the optimal global convergence rate~\cite{nesterov2013introductory}, which improves the local convergence of HB.

Nowadays, neural networks have gained remarkable performance in a various tasks owning to their over-parameterization, such as  GPT-3 with 175 billion parameters~\cite{Brown2020}.
As a result, the training process of  neural networks  suffers from  significant computation costs.
To save the expensive computational resources, momentum methods are widely adopted in the deep learning community due to their acceleration  over GD.
Besides, NAG converges faster than HB in practical neural network training ~\cite{Schmidt2021,nado2021large}.
Unfortunately, this phenomenon lacks sufficient understandings mainly due to the non-convexity faced by the optimization with neural networks.
Recently, some advances have been achieved in providing explanations for the convergence of GD in training over-parameterized neural networks. 
Later, some works extended these findings to momentum methods~\cite{wang2020provable,Bu2021,liu2022provable}.
However, their analysis only establishes convergence guarantees, but stops short of distinguishing the convergence difference between HB and NAG.
Specifically, existing works can be roughly categorized into two parts: discrete-time and continuous-time.
For the discrete-time, \cite{wang2020provable} proved that HB can reach a global minimum when training  two-layer neural networks.
Likewise, \cite{liu2022provable} established the global convergence guarantee for NAG.
These two researches only provide convergence results with specific hyper-parameters, but it is still unclear whether NAG theoretically converges faster than HB in training neural networks. 
From the continuous-time, \cite{Bu2021} exploited the connection between the limiting ordinary differential equations~(ODEs) and the momentum methods.
However, their result relies on a low-resolution ODE, which fails to differentiate the convergence behavior between HB and NAG.
As a result, there still exists a gap between practice and theory.

Motivated by the fact that the high-resolution ODEs provide more accurate characterizations than low-resolution ODEs, we study the training dynamics of momentum methods from a high-resolution perspective and establish an explicit convergence difference between HB and NAG in training neural networks.
For the architecture of the neural network, we focus on an over-parameterized two-layer neural network with ReLU activation, which is extensively studied in prior works~\cite{Du2019,wang2020provable,liu2022provable}.
Our contributions are summarized as follows:
\begin{itemize}
	\item Firstly, we derive the residual dynamics of HB and NAG by applying high-resolution ODEs and neural tangent kernel~(NTK) theory, which reveals that the original  non-convex weight dynamics corresponds to a strongly convex residual dynamics under over-parameterized assumption.
\item Secondly, with the Lyapunov-based approach, we investigate the residual dynamics via a sharper Young's inequality instead of Cauchy-Schwarz inequality.
Compared to existing works, we improve the convergence rates of HB and NAG with tighter upper bounds.
Besides, our result provides the first theoretical guarantee for the acceleration of NAG over HB, which is mainly caused by the extra gradient correction term of NAG.
	\item Finally, we validate the findings through a convergence comparison between HB and NAG under different settings of hyper-parameters on three benchmark datasets.
	The maximum distance between the training iterate and its initial value is also tested to support our theory.
\end{itemize}

\section{Related Works}
\subsection{Momentum methods}
Momentum methods apply the historical information of gradients to accelerate GD.
For the quadratic convex problem, Polyak proved HB achieves a local iteration complexity $\mathcal{O}(\sqrt{\kappa}\log(1/\epsilon))$~\cite{polyak1964some}, where $\kappa > 1$ is the condition number of the quadratic problem.
Compared to the iteration complexity $\mathcal{O}(\kappa \log(1/\epsilon))$ of GD, HB achieves an acceleration.
For $L-$smooth and $\mu-$strongly convex problems,\cite{ghadimi2015global} proved that the average iterates of HB converge to a global optimal with $\mathcal{O}(\frac{L}{\mu} \log(1/\epsilon))$ iteration complexity, similar to the convergence of GD.
Nevertheless, \cite{lessard2016analysis} discovered a counterexample that HB fails to converge in tackling some smooth and strongly convex problems. 
In contrast, NAG enjoys a global iteration complexity $\mathcal{O}(\sqrt{\kappa}\log(1/\epsilon))$ in the same setting~\cite{nesterov2013introductory}.

Meanwhile, researchers investigated the connection between discrete momentum methods and ODEs.
With the step size towards infinitesimal, \cite{Wilson2021} derived  HB and NAG share the same ODE formulation. 
\cite{su2014differential} obtained the limiting ODE for NAG with a time-varying momentum parameter.
Subsequently, \cite{Shi2021} applied dimensional analysis to establish high-resolution ODEs for momentum methods, where HB and NAG exhibit different limiting ODEs.
However, in the continuous viewpoint, \cite{Shi2021} only proved HB and NAG have  the same convergence rate $\mathcal{O}(e^{-\sqrt{\mu}t/4})$  for $L-$smooth and $\mu-$strongly convex problems.
By discretizing the Lyapunov function and the high-resolution ODEs, Shi further proved that NAG converges faster than HB in the strongly convex setting.
Additionally, \cite{Sun2020} improved above continuous-time result of NAG with a faster $\mathcal{O}(e^{-3\sqrt{\mu}t/7})$ convergence rate.
Recently, \cite{Zhang2021} further derived a shaper $\mathcal{O}(e^{-\sqrt{\mu}t/2})$ result.
\begin{table}[H]
\centering
\caption{Comparison of existing results for HB and NAG from the high-resolution view, where SC, NC and Obj represent strongly-convex, non-convex and Objective function, respectively. } 
\vspace{-0.2cm}
\begin{tabular}{l@{\hspace{0.22cm}} c@{\hspace{0.21cm}}c@{\hspace{0.21cm}}c@{\hspace{0.2cm}} c} 
    \toprule
     & Obj  & HB & NAG \\
    \midrule
    \small{\cite{Shi2021}} & \small{SC}& \small{$\mathcal{O}(e^{-\tfrac{\sqrt{\mu}t}{4}})$} & \small{$\mathcal{O}(e^{-\tfrac{\sqrt{\mu}t}{4}})$} \\
    \midrule
    \small{\cite{Sun2020}} &\small{SC}& - & \small{$\mathcal{O}(e^{-\tfrac{3\sqrt{\mu}t}{7}})$} \\
    \midrule
    \small{\cite{Zhang2021}} &\small{SC}& - & \small{$\mathcal{O}(e^{-\tfrac{\sqrt{\mu}t}{2}})$} \\    
    \midrule
    \textbf{This work} &\small{\textbf{NC}}& \small{\textbf{$\mathcal{O}(e^{-(2\!-\!\sqrt{2})\sqrt{\mu}t})$}} & \makecell{\textbf{$\mathcal{O}(e^{-\rho_{NAG}\sqrt{\mu}t})$}\\\tiny{\textbf{$2\!-\!\sqrt{2}\!<\! \rho_{NAG}\!\leq\! \tfrac{11\!-\!\sqrt{73}}{4}$}}} \\
    \midrule
\end{tabular}
\label{1}
\end{table}
\subsection{Neural Tangent Kernel}
\cite{jacot2018neural} established the connection between the optimization of infinite wide neural networks and NTK.
Later, this result was further extended in providing convergence guarantees of GD for different architectures of neural networks, including convolutional neural network~\cite{arora2019exact}, recurrent neural network~\cite{alemohammad2020recurrent}, residual network~\cite{DBLP:journals/corr/abs-2002-06262} and graph neural network~\cite{DBLP:conf/nips/DuHSPWX19}.

With the NTK theory, there is an increasing interest in explaining  the acceleration phenomenon of momentum methods in training neural networks.
From the discrete-time view, \cite{wang2020provable} established the first convergence result of HB in training two-layer ReLU neural networks and deep linear neural networks.
Besides, their result theoretically shows the acceleration guarantee of HB over GD.
In addition, \cite{liu2022provable} presented the convergence result of NAG in training two-layer neural networks.
From the continuous-time view, \cite{Bu2021} exploited the limiting ODEs in analyzing the convergence of momentum methods for training neural networks.
However, their result relies on low-resolution ODEs, which fails to capture the convergence difference between HB and NAG.
At the same setting, our work differentiates the convergence behaviors between HB and NAG through sharper inequality and analysis.
Specifically, our result shows that HB converges at a $\mathcal{O}(e^{-(2-\sqrt{2})\sqrt{\mu}t})$ convergence rate and NAG enjoys a faster $\mathcal{O}(e^{-\rho_{NAG}\sqrt{\mu}t})$ rate, where $2-\sqrt{2}< \rho_{NAG}\leq(11-\sqrt{73})/4$.
The comparison of existing results for HB and NAG from the high-resolution view can be founded in Table~\ref{1}.

\section{Preliminary}
\subsection{Problem setting}
\label{problem setting}
In this paper, we focus on the supervised learning problem for training a two-layer ReLU neural network with $m$ hidden nodes:
\begin{equation}
\label{nn}
    f(\W, \bm{a}, \x) = \frac{1}{\sqrt{m}}\sum_{r=1}^m a_r \sigma (\w_r^{\top} \x),
\end{equation}
where $\x \!\in\! \R^{d}$ denotes the input feature, $\sigma (z)\!=\!\max\{0, z\}$ denotes the ReLU activation function, $\W \!=\!(\w_1,\cdots, \w_m) \in \R^{d \times m}$ is the parameter of the hidden layer and $\bm{a} \!\in\! \R^{m}$ is the parameter of the output layer.
For simplicity, we assume the input feature has unit norm, i.e., $\|\x\|_2 \!=\! 1$, which follows~\cite{Bu2021,Du2019}.
For initialization, $\w_r(0) \!\sim\! \mathcal{N}(\bm{0}, \bm{I}_d)$ and $a_r \!\sim\! Rademaher(0.5)$ for any $r\!\in\! [m]$. 
According to~\cite{wang2020provable,Bu2021,Du2019}, we use the square loss as the loss function.
Then, it can find the optimal parameter by minimizing the empirical risk $\ell$ to :
\begin{equation}
\label{two-layer}
	\ell(\W, \bm{a}) = \frac{1}{2}\sum_{i=1}^n (f(\W, \bm{a}, \x_i) - y_i)^2,
\end{equation}
where $(\x_i, y_i) \in \R^d \times \R$ denotes the $i-$th training instance.

Following~\cite{Bu2021,Du2019}, we only update $\W$ and keep $\bm{a}$ fixed during training.
Note that the problem in Eq.(\ref{two-layer}) is both non-convex and non-smooth. 
Recently, some works  provided  global convergence guarantees for gradient-based methods in optimizing Eq.(\ref{two-layer})~\cite{Bu2021,Du2019}.
GD is the most widely used gradient-based method due to its simplicity and efficiency, which updates as
\begin{eqnarray}
	\w_r(t+1) = \w_r(t) - \eta \tfrac{\partial \ell(\W(t), \bm{a})}{\partial \w_r(t)}, 
\end{eqnarray}
where $\eta > 0$ is the learning rate.

For accelerating GD, momentum methods blend history information of gradients into current updates.
HB starts from the initial parameter $\w_r(-1) \!=\! \w_r(0)$ and updates as follows
\begin{equation}
	\label{eq:HB_update}
\w_r(t\!+\!1) \!=\! \w_r(t) \!+\! \beta\big(\w_r(t) \!-\! \w_r(t\!-\!1)\big) \!-\! \eta \tfrac{\partial \ell(\W(t), \bm{a})}{\partial \w_r(t)},
\end{equation}
where $\beta \in [0, 1)$ denotes the momentum parameter.
According to~\cite{nesterov2013introductory}, NAG has several variants with different schemes of hyper-parameters.
In this paper, we mainly focus on NAG with a constant momentum parameter $\beta$.
Given the initial parameters $\w_r(-1) = \w_r(0)$,
NAG involves the update procedure as
\begin{eqnarray}
	\label{eq:NAG-SC_one_line}
	\w_r(t\!+\!1) \!\!\!\!&=&\!\!\!\! \w_r(t)  \!+\! \beta\big(\w_r(t) \!-\! \w_r(t\!-\!1)\big) \!-\! \eta \tfrac{\partial \ell(\W(t), \bm{a})}{\partial \w_r(t)} \nonumber\\
	&&- \beta\eta\big(\tfrac{\partial \ell(\W(t), \bm{a})}{\partial \w_r(t)} \!-\! \tfrac{\partial \ell(\W(t-1), \bm{a})}{\partial \w_r(t-1)}\big).
\end{eqnarray}
Compared to HB, NAG has an extra term $\beta\eta(\frac{\partial \ell(\W(t), \bm{a})}{\partial \w_r(t)} \!-\! \frac{\partial \ell(\W(t\!-\!1), \bm{a})}{\partial \w_r(t\!-\!1)})$, which is associated with the difference between two consecutive gradients and is referred to as gradient correction~\cite{Shi2021}.
This term is crucial for the provable acceleration of NAG over HB in the convex setting, but lacks enough understandings in the non-convex setting.

\subsection{Continuous view of momentum methods}
\label{sec3.2}
\textbf{Continuous view of GD}. Before introducing the limiting ODEs of momentum methods, we present existing result of GD in training two-layer ReLU neural networks.

When learning rate tends to infinitesimal, GD corresponds to the gradient flow
\begin{eqnarray}
	\frac{d \w_r(t)}{dt} = - \frac{\partial \ell(\W(t), \bm{a})}{\partial \w_r(t)}.
\end{eqnarray}
Based on Eq.(\ref{two-layer}), the time derivative of the hidden layer has
\begin{eqnarray}
	\frac{d \w_r(t)}{dt} \!=\! - \frac{\partial \ell(\W(t), \bm{a})}{\partial \w_r(t)} \!=\! -(\frac{\partial \f(t)}{\partial \w_r(t)})^{\top}(\f(t) \!-\!\y),
\end{eqnarray}
where $\f(t)=(f_1, \cdots, f_n)^{\top}\in \R^{n}$ with $f_i = f(\W(t), \bm{a};\x_i)$ and $\y = (y_1, \cdots, y_n)^{\top} \in \R^{n}$. 
Denotes $\bm{\Delta} = \f - \y$ as the residual vector, it has
\begin{eqnarray}
	\frac{d \f(t)}{dt} = \sum_r \frac{\partial \f(t)}{\partial \w_r(t)} \frac{                                                                    d \w_r(t)}{dt} = -\bm{H}(t)(\f(t) - y),
\end{eqnarray}
where $\bm{H}(t)$ is a Gram matrix induced by the two-layer ReLU neural network with following formulation
\begin{equation}
\label{H}
	\bm{H}(t) := \sum_{r=1}^m \frac{\partial \f(t)}{\partial \w_r(t)} (\frac{\partial \f(t)}{\partial \w_r(t)})^{\top}.
\end{equation}
When $m$ goes to infinity, $\bm{H}$ at the initialization has 
\begin{eqnarray}
\label{NTK_matrix}
 \bm{H}_{ij}^{\infty} \!\!\!\!&:=&\!\!\!\! \lim_{m \to \infty} \bm{H}_{ij}(0) \nonumber\\
 &=&\!\!\!\!\!\!\mathbb{E}_{w \sim \mathcal{N}(\bm{0},\bm{I})} [\x_i^{\top}\x_j \I\{\w^{\top}\!\x_i \!\geq\! 0, \w^{\top}\!\x_j \!\geq\! 0\}]\!, 
\end{eqnarray} 
which holds for any $i, j \in [n]$.
Note that $\bm{H}^{\infty}$ is positive definite with following lemma.

\begin{lemma}[Theorem 3.1 in \cite{Du2019}]
Suppose $\x_i$ is not parallel with $\x_j$ for any $i \!\neq\!
 j$, then $\lambda_0 :=\lambda_{min}(\bm{H}^{\infty}) \!>\! 0$.
\end{lemma}

When $m$ is large enough,
$\bm{H}(0)$ is also positive definite.

\begin{lemma}[Lemma 3.1 in \cite{Du2019}]
If $m\!=\!\Omega\left(\frac{n^2}{\lambda_{0}^{2}} \log \left(\frac{n^2}{p}\right)\right)$, with probability at least $1-p$, it has $\left\|\bm{H}(0)-\bm{H}^{\infty}\right\|_{2} \leq \frac{\lambda_{0}}{4}$ and $\lambda_{\min }(\bm{H}(0)) \geq \frac{3\lambda_{0}}{4} $.
\label{lem:B1}
\end{lemma}

Next, we introduce a lemma that shows for any $t$, if $\w_r(t)$ is close to $\w_r(0)$, then $\bm{H}(t)$ is identical to $\bm{H}(0)$, which ensures the positive definite of $\bm{H}(t)$.

\begin{lemma}[Lemma 3.2 in \cite{Du2019}]
\label{lemma3.3}
Assume $\bm{w}_r(0) \sim \mathcal{N}(\bm{0},\bm{I})$ for $r \in [m]$ and $\left\|\bm{w}_{r}(0)-\bm{w}_{r}\right\|_{2} \leq \frac{c p \lambda_{0}}{n^2} =: R$ for some small positive constant $c$, with probability at least $1-p$, it has $\|\bm{H}(t)-\bm{H}(0)\|_2<\frac{\lambda_0}{4}$, $\lambda_{\min}(\bm{H}(t))>\frac{\lambda_0}{2}$ and $\lambda_{max}(\bm{H}(t)) < \lambda_m :=\lambda_{max}(\bm{H}^{\infty}) + \frac{\lambda_0}{4}$.
\label{lem:B2}
\end{lemma}

Therefore, under the over-parameterized assumption, $\bm{H}(t)$ stays positive with the condition number $\kappa := \lambda_m/(\lambda_0/2) > 1$ and $\w_r(t)$ remains close to its initialization during the training process.

\noindent\textbf{Low-resolution view of momentum methods.} From a continuous view, \cite{Bu2021} derived the convergence guarantees of HB and NAG in optimizing Eq.(\ref{two-layer}).
They consider a non-linear dissipative dynamical system with $b > 0$
\begin{equation}
\label{ode: momentum}
	\ddot{\w}_r(t) + b\dot{\w}_r(t) + \tfrac{\partial \ell(\W(t), \bm{a})}{\partial \w_r(t)} = 0,
\end{equation}
which corresponds to the limiting ODE for both HB and NAG when the learning rate tends to infinitesimal~\cite{Wilson2021}.
Consequently, Eq.(\ref{ode: momentum}) fails to capture the convergence  difference between HB and NAG.

\noindent\textbf{High-resolution view of momentum methods.}
In this subsection, we first present the high-resolution formulations of HB and NAG, then derive their corresponding residual dynamics when training the two-layer ReLU neural network.

On the one hand, according to \cite{Shi2021}, HB has the following high-resolution ODE
\begin{equation}
\label{eq:HB_W}
\ddot{\W}(t) + 2\sqrt{\alpha}\dot{\W}(t) + (1+\sqrt{ \alpha s})\tfrac{\partial \ell(\W(t), \bm{a})}{\partial \W(t)} = 0,
\end{equation}
where $\alpha \geq 0$ and $s > 0$.
For each neuron $r \in [m]$,  HB has the corresponding ODE
\begin{equation}
\label{eq:HB}
	\ddot{\w}_r(t) + 2\sqrt{\alpha}\dot{\w}_r(t) + (1+\sqrt{ \alpha s})\tfrac{\partial \ell(\W(t), \bm{a})}{\partial \w_r(t)} = 0.
\end{equation}
On the other hand, NAG has a different ODE representation
\begin{eqnarray}
\label{eq:NAG_W}
\ddot{\W}(t) &+& 2\sqrt{\alpha}\dot{\W}(t) + \sqrt{s} \tfrac{\partial^2 \ell(\W(t), a)}{\partial \W^2(t)} \dot{\W}(t) \nonumber\\
&+& (1+\sqrt{\alpha s}) \tfrac{\partial \ell(\W(t), \bm{a})}{\partial \W(t)} = 0.
\end{eqnarray}
The third term in Eq.(\ref{eq:NAG_W}) is the main difference between NAG and HB, which is induced by the gradient correction term in Eq.(\ref{eq:NAG-SC_one_line})~\cite{Shi2021}.
Accordingly, NAG has the following ODE for any $\w_r$
\begin{eqnarray}
\label{eq:NAG}
	\ddot{\w}_r(t) &+& 2\sqrt{\alpha}\dot{\w}_r(t) + \sqrt{s} \sum_{i \in [m]}\tfrac{\partial^2 \ell(\W(t), a)}{\partial \w_r(t)\partial \w_i(t)} \dot{\w}_i(t) \nonumber\\
	&+& (1+\sqrt{\alpha s}) \tfrac{\partial \ell(\W(t), \bm{a})}{\partial \w_r(t)} = 0.
\end{eqnarray}
When $s \!\to\! 0$, Eq.(\ref{eq:HB}) and Eq.(\ref{eq:NAG}) degenerate to the same low-resolution Eq.(\ref{ode: momentum}).
However, $\ell$ is non-convex while \cite{Shi2021} derived the convergence rate  only in the convex setting. 
For the $\mu-$strongly convex and smooth problem $\ell$, Shi \textit{et al.} sets $\alpha \!=\! \mu$ for both Eq.(\ref{eq:HB}) and Eq.(\ref{eq:NAG}).
However, their result only shows that HB and NAG converge at the same rate $\mathcal{O}(e^{-\sqrt{\mu}t/4})$, which fails to distinguish them in the continuous view.

Instead of the convex setting, we mainly focus on a non-convex  $\ell$.
Following~\cite{Bu2021}, it has
\begin{eqnarray}
\label{eq:16}
	\frac{\partial f_i}{\partial \w_r} \!\!\!&=&\!\!\! \frac{1}{\sqrt{m}}a_r \x_i \I\{\langle \w_r, \x_i\rangle \geq 0\},\\
	\label{eq:17}
	\frac{\partial^2 f_i}{\partial \w_r \partial \w_l} \!\!\!&=&\!\!\! \bm{0}_{d \times d} \;\; \text{for} \; r \neq l ,\\
	\label{eq:18}
	\frac{\partial^2 f_i}{\partial \w_r^2} \!\!\!&=&\!\!\! \frac{1}{\sqrt{m}}a_r \x_i\x_i^{\top} \delta(\langle \w_r, \x_i\rangle) \overset{a.s.}{=} \bm{0}_{d \times d},
\end{eqnarray}
where $\bm{0}_{d \times d}$ represents the zero matrix with dimension $d \times d$, $\frac{\partial \I\{x \geq 0\}}{\partial x} = \delta(x)$ and $\delta$ denotes the delta function that
\begin{equation}
	\delta(x) = \begin{cases}
+\infty,  & x = 0 \\
0, & x \neq 0.
\end{cases}
\end{equation}
Then, it has
\begin{eqnarray}
\label{dot_f}
	\dot{f}_i \!\!\!&=&\!\!\! \sum_r \langle \frac{\partial f_i}{\partial \w_r}, \dot{\w_r} \rangle, \;\;\;\;\\
\label{ddot_f}
	\ddot{f}_i \!\!\!&=&\!\!\! \sum_{r, l \in [m]} \dot{\w}_r^{\top} \frac{\partial^2 f_i}{\partial \w_r \partial \w_l}\dot{\w}_l + \sum_{r \in [m]} \langle \frac{\partial f_i}{\partial \w_r}, \ddot{\w}_r \rangle\nonumber\\
	& \overset{a.s.}{=}& \sum_{r\in [m]} \langle \frac{\partial f_i}{\partial \w_r}, \ddot{\w}_r \rangle,
\end{eqnarray}
and
\begin{eqnarray}
\label{second_order}
	\frac{\partial^2 \ell }{\partial \w_r^2} \!\!\!&=&\!\!\! \left( \frac{\partial \f}{\partial \w_r}\right)^{\top}\frac{\partial \f}{\partial \w_r} + \frac{\partial^2 \f}{\partial \w_r^2}(\f-\y)\nonumber \\
	&\overset{a.s.}{=}&  \left( \frac{\partial \f}{\partial \w_r}\right)^{\top}\frac{\partial \f}{\partial \w_r},\\
\label{second_order_diff}
\frac{\partial^2 \ell }{\partial \w_r \partial \w_l} \!\!\!&=&\!\!\! \left( \frac{\partial \f}{\partial \w_r}\right)^{\top}\frac{\partial \f}{\partial \w_l} + \frac{\partial^2 \f}{\partial \w_r \partial \w_l}(\f-\y)\nonumber\\
& {=}&  \left( \frac{\partial \f}{\partial \w_r}\right)^{\top}\frac{\partial \f}{\partial \w_l} \;\; \text{for} \; r \neq l .
\end{eqnarray}
When multiplying Eq.(\ref{eq:HB}) with $\frac{\partial \f}{\partial \w_r}$ and summing over all $r\!\in\![m]$, it has the following ODE based on Eq.(\ref{H}) and Eq.(\ref{dot_f})
\begin{equation}
	\label{eq:HB_predy}
	\ddot{\f}(t) + 2\sqrt{\alpha}\dot{\f}(t) + (1+\sqrt{\alpha s})\bm{H}(t)(\f(t)-\y)=0.
\end{equation}
Then, the dynamics of the residual vector $\Delta = \f - \y$ has
\begin{equation}
	\label{eq:HB_predy_err}
	\ddot{\Delta}(t) + 2\sqrt{\alpha}\dot{\Delta}(t)+(1+\sqrt{\alpha s})\bm{H}(t)\Delta(t) = 0.
\end{equation}

Instead of directly analyzing the non-convex $\ell$ over $\W$, we show that optimizing $\ell$ corresponds to  the pseudo-loss $\hat{\ell}(t):=\frac{1}{2}\Delta^{\top}(t)\bm{H}(t) \Delta(t)$, which is $\frac{\lambda_0}{2}-$strongly convex over $\Delta(t)$ according to Lemma~\ref{lem:B2}.
With $\mu = \lambda_0/2$, Eq.(\ref{eq:HB}) can be transformed to
\begin{equation}
	\label{eq:HB_error}
	\ddot{\Delta}(t) + \sqrt{2\lambda_0}\dot{\Delta}(t)+(1+\sqrt{\tfrac{\lambda_0 s}{2}})\tfrac{\partial \hat{\ell}(t)}{\partial \Delta(t)} = 0,
\end{equation}
where $\alpha = \lambda_0/{2}$ according to \cite{Shi2021}.
Similarly, it can derive following residual dynamics of NAG as
\begin{eqnarray}
\label{eq:NAG_error}
\ddot{\Delta}(t) \!\!+\!\! \sqrt{2\lambda_0}\dot{\Delta}(t) \!\!+\!\! \sqrt{s}\bm{H}\!(t)\dot{\Delta}\!(t) \!\!+\!\! (1\!\!+\!\!\sqrt{\tfrac{\lambda_0 s}{2}})\tfrac{\partial \hat{\ell}(t)}{\partial \Delta(t)} \!\!=\!\! 0,
\end{eqnarray}
where the third term in Eq.(\ref{eq:NAG_error}) is derived by
\begin{eqnarray}
&&\sum_{r \in [m]}\big\langle \frac{\partial \f}{\partial \w_r}, \sum_{i \in [m]}\frac{\partial^2 \ell(\W(t), a)}{\partial \w_r(t)\partial \w_i(t)} \dot{\w_i}(t) \big\rangle \nonumber\\
&=&\!\!\!\! \sum_{r \in [m]} \big\langle \frac{\partial \f}{\partial \w_r}, \sum_{i \in [m]}\left( \frac{\partial \f}{\partial \w_r}\right)^{\top}\frac{\partial \f}{\partial \w_i} \dot{\w_i}(t)\big\rangle \nonumber\\
&=&\!\!\!\!\!\! \big\langle \sum_{r\in [m]} \frac{\partial \f}{\partial \w_r}\left(\frac{\partial \f}{\partial \w_r}\right)^{\top}, \sum_{i \in [m]} \frac{\partial \f}{\partial \w_i} \dot{\w_i}(t) \big\rangle =\bm{H}(t)\dot{\f}(t).\nonumber
\end{eqnarray}
Note that the third term in Eq.(\ref{eq:NAG_error}) is derived from the gradient correction term.

\section{Main results}
\label{main results}
In this section, we provide the convergence analysis and comparison of HB and NAG.
\subsection{Convergence analysis of HB}

\begin{thm}
\label{thm_HB}
 Suppose $m = \Omega(\frac{n^6}{p^3\lambda_0^4})$, $\w_r(0) \sim \mathcal{N}(\bm{0},\bm{I}_d)$ and $a_r \sim unif\{-1, 1\}$ for any $r \in [m]$, with probability at least $1-p$, it has	
\begin{equation}
	\ell(t) \leq \frac{6\hat{\ell}(0)}{\lambda_0} e^{-(2-\sqrt{2})\sqrt{\frac{\lambda_0}{2}} t}.
\end{equation}
\end{thm}

\noindent\textbf{Remark 1.} The above theorem implies HB can attain a global minimum of Eq.(\ref{two-layer}) at a linear rate $\mathcal{O}(e^{-(2-\sqrt{2})\sqrt{\frac{\lambda_0}{2}} t})$ in training two-layer ReLU neural networks.
It is worth noting that \cite{Shi2021} studied the same high-resolution ODE of HB and proved HB can attain a global convergence rate $\mathcal{O}(e^{-\frac{1}{4}\sqrt{\mu}t})$ for $\mu$-strongly convex problems.
In contrast, our focus is on a non-convex optimization problem.
We show that optimizing the non-convex loss $\ell$ corresponds to an error dynamics, as if we were actually optimizing  a strongly convex loss $\hat{\ell}$ over the prediction error.
Compared to~\cite{Shi2021}, Theorem~\ref{thm_HB} further provides a tighter upper bound for the convergence rate of HB in the high-resolution continuous-time view. 

\noindent\textbf{Remark 2.} 
We are aware of some recent works that improve the requirement of the over-parameterization~\cite{Song2019,Nguyen2021}. 
In this work, we mainly focus on providing the theoretical guarantee for the acceleration of NAG over HB in training over-parameterized neural networks.
Therefore, we incorporate the first and also the most well-known result about the requirement on $m$ that ensures $\bm{H}(t)$ positive~\cite{Du2019}.   
We believe incorporating these improved results on $m$ into our analysis can also improve the bound on $m$ in our work.

The proof of Theorem~\ref{thm_HB} mainly consists of three parts:
\begin{enumerate}
	\item Firstly, based on the positive definite assumption of $\bm{H}$, we  demonstrate the training loss decays to zero at a linear rate and the distance between $\w$ and its initial value has a upper bound $R^{'}$.
	\item Secondly, we show that if $R^{'}$ is smaller than the distance $R$ defined in Lemma~\ref{lemma3.3}, our conclusions from the first procedures hold true and the positive definite of $\bm{H}$ is guaranteed.
	\item Finally, the  width $m$ is determined by satisfying $R^{'} < R$, which completes the proof of Theorem~\ref{thm_HB}.
\end{enumerate}

To start with, we present the first procedure of the proof, which involves the smallest eigenvalue $\frac{\lambda_0}{2} $ of $\bm{H}$ as introduced in Lemma~\ref{lemma3.3}. 

\begin{lemma}
\label{lemma:4.1}
Assume $\lambda_{min}(\bm{H}(i)) \!\geq\! \frac{\lambda_0}{2}$ for $0 \!\!\leq\!\! i \!\!\leq\!\! t$.
With $0\!\!<\!\!s \!\!\leq\!\! 1/\lambda_m$ and $\dot{\w}_r(0)\!\!=\!\!0$ for any $r\!\in\! [m]$, HB has $\ell(t) \!\leq\! \frac{6\hat{\ell}(0)}{\lambda_0} e^{-(2-\sqrt{2})\sqrt{\frac{\lambda_0}{2}} t}$ and $\|\w_r(t) \!-\! \w_r(0) \|_2 \!\leq\! R^{'} \!:=\! 10\sqrt{{6\hat{\ell}(0) n}/(\lambda^3_0 m)} $.
\end{lemma}

The proof is provided in the supplementary materials. 
Then, the conclusions in Lemma~\ref{lemma:4.1} hold when the distance $R^{'}$ defined in Lemma~\ref{lemma:4.1} is less than the distance $R$  given in Lemma~\ref{lemma3.3}.

\begin{lemma}
\label{lemma:4.2}
Assume $R^{'} < R$, it has (1) $\lambda_{min}(\bm{H}(t)) \geq \lambda_0/2$, (2) $\|\w_r(t) - \w_r(0)\|_2 \leq R^{'}$ for all $r \in [m]$ and (3) $\ell(t) \leq \frac{6\hat{\ell}(0)}{\lambda_0} e^{-(2-\sqrt{2})\sqrt{\frac{\lambda_0}{2}} t}$.
\end{lemma}

The proof can be founded in the supplementary materials.	
To satisfy the assumption $R^{'} < R$ in Lemma~\ref{lemma:4.2}, 
it requires  $m=\Omega(\frac{n^6}{p^3\lambda_0^4})$ based on $\hat{\ell}(0) = \Omega(n\lambda_0/p)$ as proved in~\cite{Bu2021}, which completes the proof of Theorem~\ref{thm_HB}.

\subsection{Convergence analysis of NAG}
In this subsection, based on the similar analysis framework as the proof of HB,  we  present the convergence guarantee of NAG and show its acceleration.

\begin{thm}
\label{thm_NAG}
 Suppose $m = \Omega(\frac{n^6}{p^3\lambda_0^4})$, $\w_r(0) \sim \mathcal{N}(0, 1)$ and $a_r \sim unif\{-1, 1\}$ for any $r \in [m]$, with probability at least $1-p$, it has	
\begin{equation}
	\ell(t) \leq \frac{26\hat{\ell}(0)}{3\lambda_0} e^{-\rho_{NAG}^*(\alpha)\sqrt{\frac{\lambda_0}{2}}  t},
\end{equation}
\end{thm}
\noindent where
\begin{eqnarray}
\alpha = \sqrt{2\lambda_0 s}/4, \;\;
\rho_{NAG}^*(\alpha) = \tfrac{1}{2}(4\!+\!3\alpha \!-\! \sqrt{8\!+\!16\alpha \!+\! 9\alpha^2}).\nonumber
\end{eqnarray}

\noindent\textbf{Remark 3}. The above theorem indicates that NAG can reach a global minimum at a linear rate $\mathcal{O}(e^{-\rho_{NAG}^*(\alpha)\sqrt{\frac{\lambda_0}{2}}  t})$.
When $0< s \leq 1/\lambda_m$, it holds that $0<\alpha\leq 1/(2\sqrt{\kappa}) \leq 0.5$.
\begin{figure}[!t]
	\label{NAG_con}
	\centering
	\includegraphics[width=0.44\textwidth , height = 0.2\textwidth]{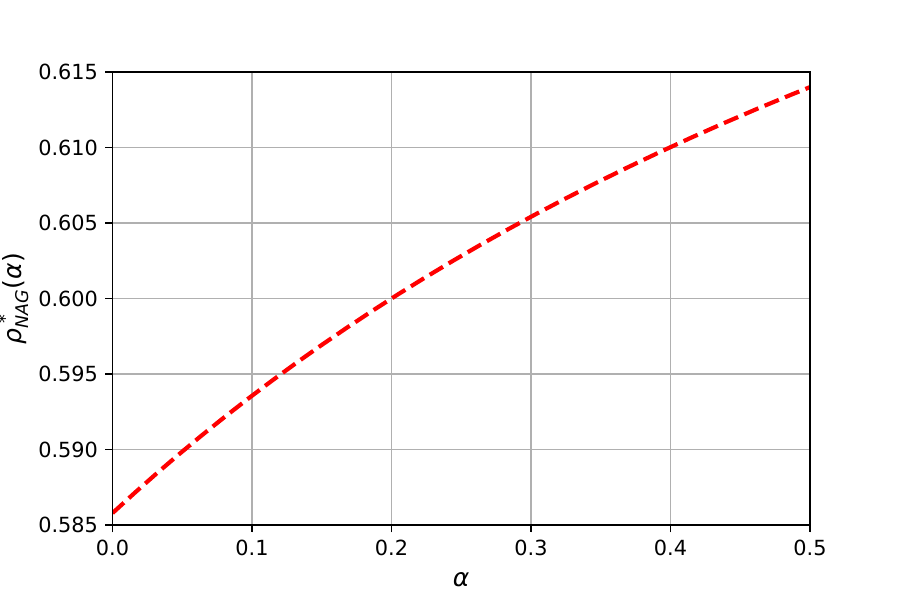}       
	\caption{The value of $\rho_{NAG}^*(\alpha)$ with respect to $\alpha$.}
	\label{fig1}
\end{figure}
From Fig.~\ref{fig1}, it can observe that $\rho_{NAG}^*(\alpha)$ is a monotonic increasing function with respect to $\alpha$ when $0< \alpha \leq 0.5$.
Thus, it has
\begin{equation}
	2 - \sqrt{2} < \rho_{NAG}^*(\alpha) \leq (11-\sqrt{73})/4.
\end{equation}
Compared to the convergence rate $\mathcal{O}(e^{-(2-\sqrt{2})\sqrt{\frac{\lambda_0}{2}} t})$ of HB, 
our theoretical result demonstrates that NAG converges faster than HB.

In addition, according to Eq.(\ref{C.1}) in the supplementary materials, the derivative of the corresponding Lyapunov function $V$ of NAG has
\begin{eqnarray}
\label{eq:31}
	\dot{V}\!\! = \!\!\!\!\!\!&-&\!\!\!\!\!\!(1\!+\!\sqrt{\tfrac{\lambda_0 s}{2}})\sqrt{2\lambda_0}\hat{\ell}(t) 
	\,\!-\! \tfrac{\sqrt{s}}{2}(1\!+\!\sqrt{\tfrac{\lambda_0 s}{2}})\|\bm{H}(t)\Delta(t)\|^2 \nonumber\\
	&-&\!\!\!\!\!\tfrac{\sqrt{2\lambda_0}}{2}\|\dot{\Delta}(t)\|^2 -\tfrac{\sqrt{s}}{2} \dot{\Delta}^{\top}(t)\bm{H}(t)\dot{\Delta}(t),
\end{eqnarray}
where the third term on the right hand side of Eq.(\ref{eq:31}) is induced from $\sqrt{s}\bm{H}(t)\dot{\Delta}(t)$, which corresponds to the gradient correction term as mentioned in Section~\ref{sec3.2}.
\begin{figure*}[!htb]
\centering
\includegraphics[width=0.97\textwidth]{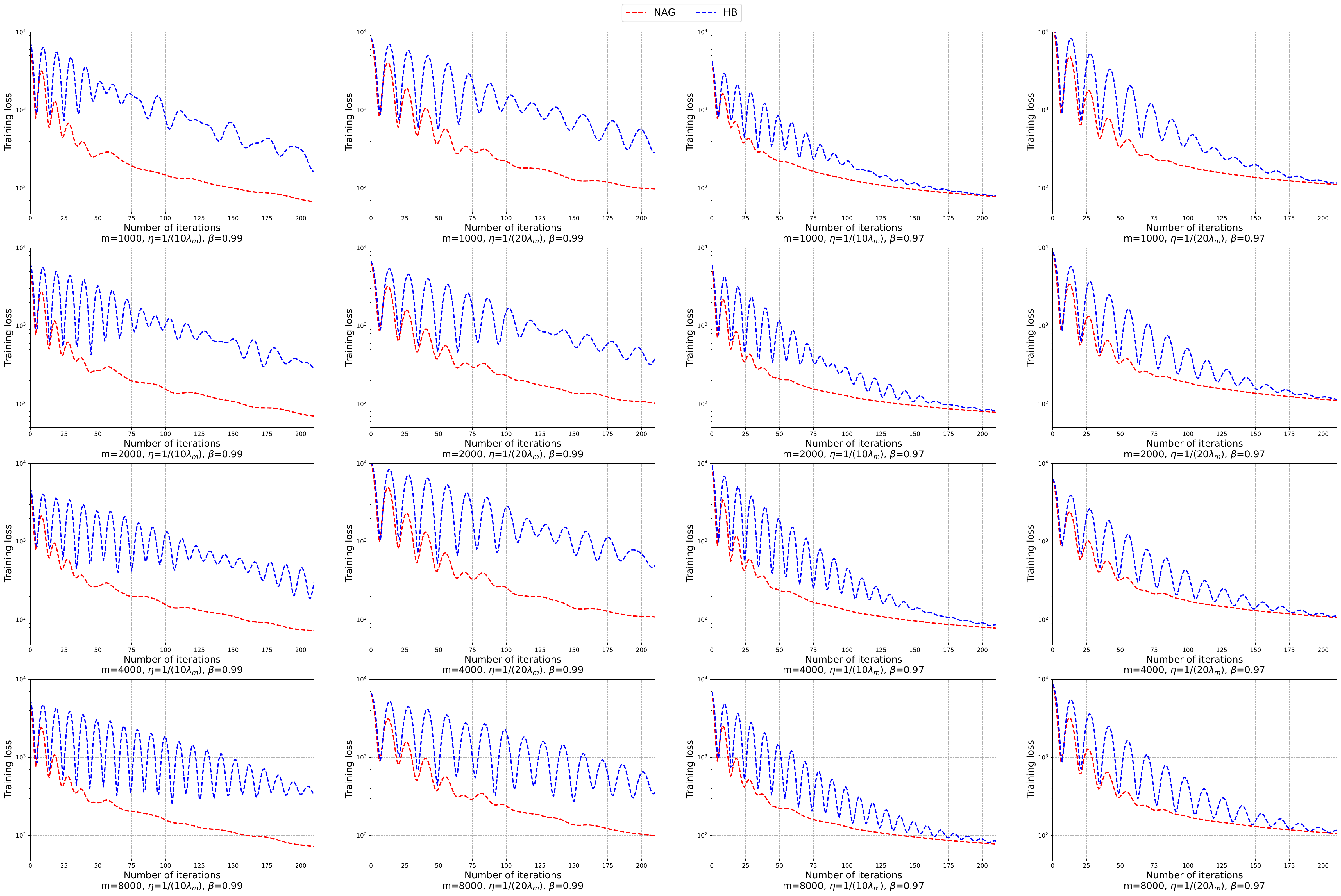} 
\caption{Convergence comparison between NAG and HB in training two-layer ReLU neural networks on the MNIST dataset with different width $m$, learning rate $\eta$ and momentum parameter $\beta$.}
\label{fig2}
\end{figure*}
If neglecting this term, based on the upper bound of $V(t)$ (Eq.(\ref{lemma6_V}) in the supplementary materials), the corresponding convergence rate has
\begin{eqnarray}
\rho^* = \max_{\phi>0} \min\{ \tfrac{2}{3+2\phi}, \tfrac{4}{2+1/\phi}, \tfrac{4}{1+\phi}\} \sqrt{\tfrac{\lambda_0}{2}} 
= 2 - \sqrt{2}, \nonumber
\end{eqnarray}
which is the same as the convergence rate of HB.
Therefore, we provably show that the extra gradient correction term of NAG in Eq.(\ref{eq:NAG_error}) leads to the acceleration of NAG over HB in training neural networks.

\noindent\textbf{Remark 4.} 
It is noted that \cite{Shi2021} conducted an analysis on the same high-resolution ODE of NAG for $\mu$-strongly convex problems.
Their result demonstrated that the convergence rate of NAG is upper bounded by $\mathcal{O}(e^{-\frac{1}{4}\sqrt{\mu} t})$.
Later, \cite{Sun2020} and \cite{Zhang2021} improved the convergence rate to $\mathcal{O}(e^{-\frac{3}{7}\sqrt{\mu} t})$ and $\mathcal{O}(e^{-\frac{1}{2}\sqrt{\mu} t})$, respectively.
Comparing to~\cite{Shi2021,Sun2020,Zhang2021}, our study focuses on a non-convex $\ell$ over the weights.
Furthermore, we demonstrate that $\ell$ corresponds to a $\frac{\lambda_0}{2}$-strongly convex problem by utilizing the NTK theory.
Additionally, we provide a tight convergence rate of NAG with a more refined inequality and analysis.

\noindent\textbf{Remark 5.} Noted that directly studying the convergence of discrete algorithms for non-convex $\ell$ is far too complex and general to offer insightful explanation about their efficiency in deep learning. 
Instead of the low-resolution approach in previous work~\cite{Bu2021}, we study from a high-resolution continuous viewpoint.
Our result provides the first acceleration guarantee of NAG over HB with a simple and intuitive proof.
Furthermore, Proposition 1 in~\cite{Shi2021} shows that the discrete iterates and the high-resolution ODE trajectories remain close for both HB and NAG when the learning rate is small enough. 
It would be interesting to consider the discrete setting in the future work.

\begin{figure*}[!h]
\centering
\includegraphics[width=0.97\textwidth , height=0.2\textwidth]{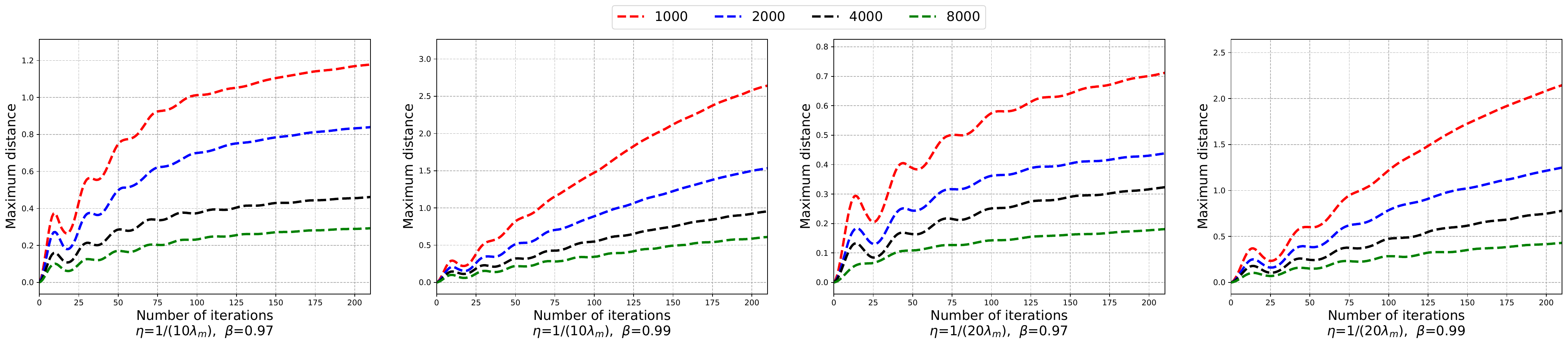}
\caption{The maximum distance $\max_{r \in [m]}\|\w_r(t) - \w_0(t)\|$ of HB in training two-layer ReLU neural networks on the MNIST dataset during training process.  }
\label{fig3}
\end{figure*}

\begin{figure*}[!h]
\centering
\includegraphics[width=0.97\textwidth , height=0.2\textwidth]{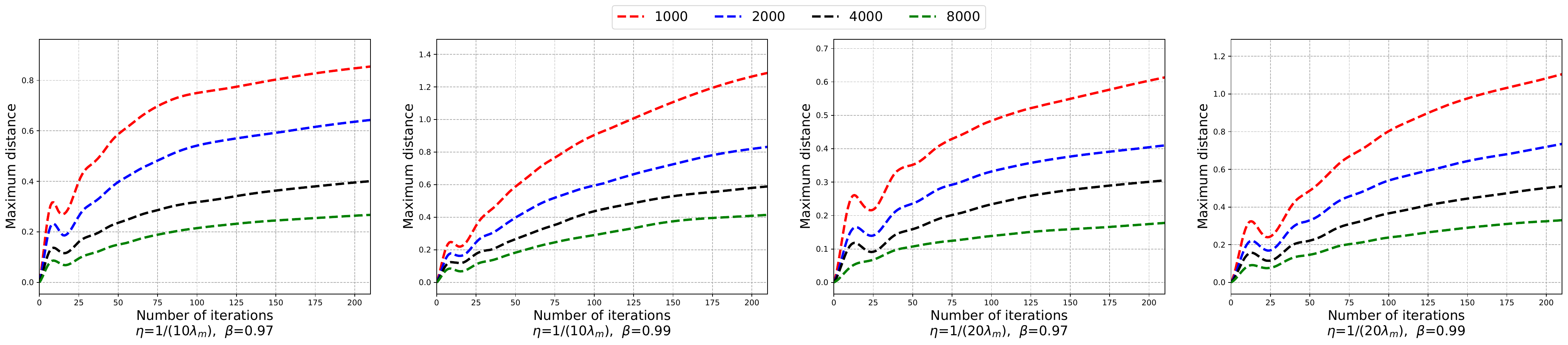} 
\caption{The maximum distance $\max_{r \in [m]}\|\w_r(t) - \w_0(t)\|$ of NAG in training two-layer ReLU neural networks on the MNIST dataset during training process.}
\label{fig4}
\end{figure*}

\begin{lemma}
\label{lemma:4.3}
Assume $\lambda_{min}(\bm{H}(i)) \geq \frac{\lambda_0}{2}$ for $0 \leq i \leq t$.
With $0<s \leq 1/\lambda_m$ and $\dot{\w}_r(0) = 0$ for any $r\in[m]$, NAG has $\ell(t) \leq \frac{26\hat{\ell}(0)}{3\lambda_0} e^{-\rho_{NAG}^*(\alpha)\sqrt{\lambda_0/2} t}$ and $\|\w_r(t) - \w_r(0) \| \leq 67\sqrt{\hat{\ell}(0) n/(\lambda^3_0 m)}$.
\end{lemma}

The proof is presented in the supplementary materials.
Similarly, it can establish the conclusion of the Theorem~\ref{thm_NAG} following Lemma~\ref{lemma:4.2}.
Moreover, the required width should meet $m = \Omega\big(\max\{{\sqrt{s}n/\sqrt{\lambda_0}}, \frac{n^6}{p^3\lambda_0^4}\}\big) = \Omega(\frac{n^6}{p^3\lambda_0^4})$ using $\lambda_0 \leq 1/2$ as proved in~\cite{Bu2021}.

\section{Numerical experiments}
\subsection{Setup}
We use three commonly used datasets: MNIST~\cite{lecun1998gradient}, FMNIST~\cite{xiao2017fashion} and CIFAR10~\cite{krizhevsky2009learning}, where
the pre-processing procedure follows the existing work~\cite{ADHLSW19_icml}.
Specifically, we use the first two classes of instances,  assigning the label of the first class to +1 and the second to -1.
In addition, each instance is normalized with the unit norm.

Regarding the architecture and initialization of the neural network, we adopt the setting in Section~\ref{problem setting}.
Besides, we set the width $m \in \{1000, 2000, 4000, 8000\}$,  learning rate $\eta \in \{1/(10\lambda_m), 1/(20\lambda_m)\}$ and momentum parameter $\beta \in \{0.97, 0.99\}$ , where $\lambda_m$ is calculated with the eigenvalues of the NTK matrix in Eq.(\ref{NTK_matrix}).
For each setting of hyper-parameters, we conduct 5 independent runs with different random seeds for initialization.
The average training loss over iterations is presented by the dashed line.
We use JAX~\cite{jax2018github} library for code implementation and all experiments are conducted on 4 NVIDIA V100 GPU.

\subsection{Results analysis}
Due to the space limitation, only the results on the MNIST dataset are presented here.
The remaining results can be founded in the supplementary materials.
From Fig.~\ref{fig2},  it can observe that the training losses of NAG and HB decay at a almost linear rate, which demonstrates the linear convergence of HB and NAG in training the two-layer ReLU neural networks.
Additionally,  the loss curve of NAG lies below that of HB for different hyper-parameter settings, which indicates NAG achieves a smaller training loss compared to HB at the same iteration.
Thus, NAG achieve an acceleration over HB.
These empirical findings validate the conclusion in Section~\ref{main results} that NAG theoretically obtains a faster global convergence rate compared to HB.

To further validate our findings, we present the maximum distance $\max_{r \in [m]}\|\w_r(t) \!-\! \w_0(t)\|$ of HB and NAG during the training process with different widths.
According to Lemma~\ref{lemma:4.1} and Lemma~\ref{lemma:4.3}, the theoretical maximum distances of HB and NAG both satisfy $\max_{r \in [m]}\|\w_r(t) \!-\! \w_0(t)\| \sim \mathcal{O}(1/\sqrt{m})$, which indicate that the maximum distance decreases when the width of the neural network increases.
The corresponding empirical results  are shown in Fig.~\ref{fig3} and Fig.~\ref{fig4}, which is in accordance with our theoretical findings.

\section{Conclusion and Future Work}
In this paper, we study the convergence of momentum methods in training an over-parameterized two-layer ReLU neural network.
By exploiting high-resolution ODE formulations and the NTK theory, the global convergences of HB and NAG are established.
In contrast with previous works, we exploit high-resolution ODEs instead of the low-resolution.
In addition, our work yields tighter upper bounds via a more refined inequality and analysis, which provides the first theoretical guarantee for the acceleration of NAG over HB in tackling non-convex optimization problem on neural networks.

There are several directions for future research. 
Firstly, it is natural to further extend our result to deep neural networks, where the key technical challenge lies in deriving and analyzing the spectral properties of the corresponding $\bm{H}$.
We note that there is some works about the eigenvalues of $\bm{H}$, including deep fully-connected neural networks~\cite{dugradient}, graph neural networks~\cite{DBLP:conf/nips/DuHSPWX19} and convolutional neural networks~\cite{Li2019}.
Secondly, stochastic optimization methods are widely used in training neural networks due to relatively low computational costs.
Therefore, it would be interesting to extend existing stochastic convex works~\cite{Nakakita2022,Laborde2020} to analyze the stochastic optimization problem in neural networks.
Finally, the understanding of the adaptive first-order method is still lacking, such as~AdaBound~\cite{LiuKXQL22} and RMSProp~\cite{XuZZM21}.
These adaptive methods also incorporate the idea of momentum and
our work may inspire the development of new tools for understanding their convergence behaviors.

\section*{Acknowledgement}
{This work was supported by National Natural Science Foundation of China (No.62076251 and No.62106281).
 
\bibliographystyle{named}
\bibliography{High}
\onecolumn
\appendix
\section{Supplementary materials}
\begin{enumerate}\setlength\itemsep{0em}
\renewcommand{\labelenumi}{A.\theenumi}
  \item \hyperref[proof:4.1]{Proof of Lemma 4}
  \item \hyperref[proof4.2]{Proof of Lemma 5}
  \item \hyperref[proof4.3]{Proof of Lemma 6}
  \item \hyperref[additional experiments]{Additional experiments}
\end{enumerate}
\subsection{Proof of Lemma~\ref{lemma:4.1}}
\label{proof:4.1}
\begin{proof}
	Inspired by~\cite{Shi2021,Sun2020}, we use the Lyapunov function
\begin{eqnarray}
	V(t) := (1+\sqrt{\frac{\lambda_0 s}{2}})\hat{\ell}(t) + \frac{1}{4}\|\dot{\Delta}(t)\|_2^2 + \frac{1}{4}\|\dot{\Delta}(t)+\sqrt{2\lambda_0}\Delta(t)\|_2^2.
\end{eqnarray}
Instead of Cauchy-Schwarz inequality used in~\cite{Shi2021}, we use a refined inequality: Young's inequality.
It has
\begin{equation}
	V(t) \leq (1+\sqrt{\frac{\lambda_0 s}{2}})\hat{\ell}(t) + \frac{1}{4}(2+\frac{1}{\phi})\|\dot{\Delta}(t)\|_2^2 + \frac{(1+\phi)\lambda_0}{2} \|\Delta(t)\|_2^2 \nonumber,
\end{equation}
where $\phi > 0$.

In addition, the derivative of $V(t)$ has the bound
\begin{eqnarray}
	\label{HB:derivate_lya}
	\dot{V}(t) &= &(1+\sqrt{\frac{\lambda_0 s}{2}})(\dot{\Delta}(t)^{\top}H(t)\Delta(t) + \frac{1}{2} \Delta(t)^{\top}\dot{H}(t)\Delta(t)) + \frac{1}{2}\langle \dot{\Delta}(t),  \ddot{\Delta}(t) \rangle \nonumber\\
	& &+  \frac{1}{2}\langle \dot{\Delta}(t)+\sqrt{2\lambda_0}\Delta(t) , \ddot{\Delta}(t)+\sqrt{2\lambda_0}\dot{\Delta}(t)  \rangle \nonumber\\
	&\overset{a}{=}& (1+\sqrt{\frac{\lambda_0 s}{2}})\dot{\Delta}(t)^{\top}H(t)\Delta(t) + \frac{1}{2}\langle \dot{\Delta}(t),  -\sqrt{2\lambda_0}\dot{\Delta}(t)-(1+\sqrt{\frac{\lambda_0 s}{2}})H(t)\Delta(t) \rangle \nonumber\\
	&& + \frac{1}{2}\langle \dot{\Delta}(t)+\sqrt{2\lambda_0}\Delta(t), -(1+\sqrt{\frac{\lambda_0 s}{2}})H(t)\Delta(t) \rangle \nonumber \\
	&\overset{}{=}& -\sqrt{\frac{\lambda_0}{2}}\bigg(\|\dot{\Delta}(t)\|^2 + (1+\sqrt{\frac{\lambda_0 s}{2}})\langle H(t)\Delta(t), \Delta(t)  \rangle \bigg) \nonumber\\
	&\overset{b}{\leq}& -\sqrt{\frac{\lambda_0}{2}}\bigg((1+\sqrt{\frac{\lambda_0 s}{2}})2z\hat{\ell}(t) + \|\dot{\Delta}(t)\|_2^2 +(1-z){\frac{\lambda_0}{2}}\|\Delta(t)\|_2^2\bigg),\nonumber
\end{eqnarray}
where (a) uses $\dot{H}(t) \overset{a.s.}{=}  0$ according to~\cite{Bu2021}, (b) uses $\Delta^{\top}(t)H(t)\Delta(t)\geq \frac{\lambda_0}{2}\|\Delta(t)\|_2^2$ and $0\leq z \leq 1$.
Therefore, with the bounds of $V(t)$ and $\dot{V}(t)$, HB has
\begin{equation}
\label{convergence_rate_HB}
\rho_{HB}^* = \max_{\phi>0, 0\leq z\leq 1} \min\{2z, \frac{4}{2+\frac{1}{\phi}}, \frac{1-z}{1+\phi}\}\sqrt{\frac{\lambda_0}{2}} ,
\end{equation}
According to the relationship between $2z, \frac{4}{2+\frac{1}{\phi}}$, and $\frac{1-z}{1+\phi}$, it has
\begin{equation}
	\min\{2z, \frac{4}{2+\frac{1}{\phi}}, \frac{1-z}{1+\phi}\} = \begin{cases}
2z,  & 2z \leq \frac{4}{2+1/\phi}, 2z \leq \frac{1-z}{1+\phi} \\
\frac{1-z}{1+\phi}, & 2z \leq \frac{4}{2+1/\phi}, 2z \geq \frac{1-z}{1+\phi} \\
\frac{4}{2+1/\phi}, &  2z \geq \frac{4}{2+1/\phi} , \frac{4}{2+1/\phi} \leq \frac{1-z}{1+\phi} \\
\frac{1-z}{1+\phi}, &  2z \geq \frac{4}{2+1/\phi} , \frac{4}{2+1/\phi} \geq \frac{1-z}{1+\phi}.
\end{cases}
\end{equation}
With the constraints $\phi > 0$ and $0\leq z \leq 1$, we can derive the maximum value for each cases.
For the first case, it has the feasible region as depicted in Fig.~\ref{feasible}.
\begin{figure}[H]	
	\centering
	\includegraphics[scale=0.65]{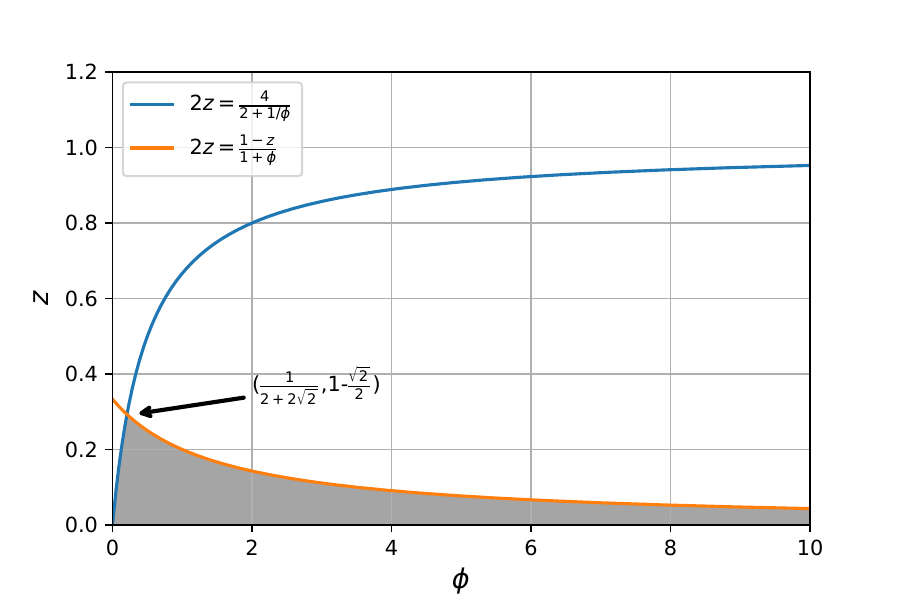}       
	\caption{The feasible region (grey area) of case 1.}
	\label{feasible}
\end{figure}
Then,  it can derive the the corresponding maximum value of $2z$ at point $(\frac{1}{2+2\sqrt{2}}, 1-\frac{\sqrt{2}}{2})$ with value $2-\sqrt{2}$. 
Similarly, it can obtain the maximum values for other three cases.
Finally, it has
\begin{equation}
\label{convergence_rate_HB_final}
\rho_{HB}^* = \max_{\phi>0, 0\leq z\leq 1} \min\{2z, \frac{4}{2+\frac{1}{\phi}}, \frac{1-z}{1+\phi}\}\sqrt{\frac{\lambda_0}{2}} =(2-\sqrt{2})\sqrt{\frac{\lambda_0}{2}} = (2-\sqrt{2})\sqrt{\frac{\lambda_0}{2}},
\end{equation}
which results in
\begin{equation}
	\dot{V}(t) \leq -\rho_{HB}^* V(t).
\end{equation}
Applying Gronwall’s inequality, it has
\begin{eqnarray}
\label{HB: ode_result}
V(t) \leq e^{-\rho_{HB}^* t} V(0).
\end{eqnarray}

Expanding $V$, it is easy to see that
\begin{eqnarray}
	(1+\sqrt{\frac{\lambda_0 s}{2}})\hat{\ell}(t) \leq V(t) \leq e^{-\rho_{HB}^* t}\big( (1+\sqrt{\frac{\lambda_0 s}{2}})\hat{\ell}(0) + \frac{1}{4}\|\dot{\Delta}(0)\|^2 + \frac{1}{4}\|\dot{\Delta}(0)+\sqrt{2\lambda_0}\Delta(0)\|^2 \big). \nonumber 
\end{eqnarray}
Based on the initial value $\dot{w}_r(0)= 0$ for any $r \in [m]$, it has
\begin{eqnarray}
	\hat{\ell}(t) &\leq& e^{-\rho_{HB}^* t}  \frac{3+\sqrt{\frac{\lambda_0 s}{2}}}{1+\sqrt{\frac{\lambda_0 s}{2}}} \hat{\ell}(0), \nonumber \\
	\ell(t) &\leq& \frac{6\hat{\ell}(0)}{\lambda_0} e^{-\rho_{HB}^* t},
\end{eqnarray}
where the first inequality exploits $\hat{\ell}(0) \geq \frac{3\lambda_0}{8} \|\Delta(0)\|_2^2$, $\dot{\Delta}(0) = 0$ and the second inequality uses $\hat{\ell}(t) \geq \frac{\lambda_0}{2} \ell(t)$.

Then, we turn to prove the bound of the distance between  $\w_r(t)$ and $\w_r(0)$.
Based on Eq.(\ref{eq:HB}), it obtains
\begin{eqnarray}
\label{formula_32}
	\frac{d}{dt}(e^{\sqrt{2\lambda_0}t}\dot{\w}_r) = -e^{\sqrt{2\lambda_0}t}(1+\sqrt{\frac{\lambda_0 s}{2}})\frac{a_r}{\sqrt{m}}\sum_{i=1}^n (f_i - y_i)\x_i\mathbb{I}\{\w_r^{\top} \x_i \geq 0\}.
\end{eqnarray}
By integrating both sides of Eq.(\ref{formula_32}), it can get 
\begin{eqnarray}
\label{formula_33}
	\dot{\w}_r(t) = -e^{-\sqrt{2\lambda_0}t} \int_0^t e^{\sqrt{2\lambda_0}t^{'}}(1+\sqrt{\frac{\lambda_0 s}{2}})\frac{a_r}{\sqrt{m}}\sum_{i=1}^n (f_i(t^{'}) - y_i)\x_i\mathbb{I}\{\w_r^{\top}\x_i \geq 0\}dt^{'}.
\end{eqnarray}
Taking the norm of Eq.(\ref{formula_33}) and applying $\sum_{i=1}^n \|z_i\|_2 \leq \sqrt{n} \|\bm{z}\|_2$, it has
\begin{eqnarray}
\label{distance_HB_new}
	\|\dot{\w}_r(t)\| &\leq& (1+\sqrt{\frac{\lambda_0 s}{2}})\frac{e^{-\sqrt{2\lambda_0}t}\sqrt{n}}{\sqrt{m}}\int_0^t e^{\sqrt{2\lambda_0}t^{'}} \|\f(t^{'}) - \y\|_2 dt^{'} \nonumber\\
	&\leq& (1+\sqrt{\frac{\lambda_0 s}{2}})\sqrt{\frac{12\hat{\ell}(0) n}{\lambda_0 m}} e^{-\sqrt{2\lambda_0}t}\frac{e^{(\sqrt{2\lambda_0} - \frac{\rho^*}{2})t} - 1}{\sqrt{2\lambda_0} - \frac{\rho_{HB}^*}{2}} \nonumber\\
&\leq& \sqrt{\frac{24\hat{\ell}(0) n}{\lambda^2_0 m}}e^{-\frac{\sqrt{2}-1}{2}\sqrt{\lambda_0}t}.
\end{eqnarray}
Applying Cauchy-Schwarz inequality on Eq.(\ref{distance_HB_new}), it has
\begin{eqnarray}
	\|\w_r(t) - \w_r(0)\|_2 \leq \int_0^t \|\dot{\w}_r(t^{'})\|_2 d t^{'} \leq 10\sqrt{\frac{6\hat{\ell}(0) n}{\lambda^3_0 m}}. \nonumber
\end{eqnarray}
\end{proof}

\subsection{Proof of Lemma~\ref{lemma:4.2}}
\label{proof4.2}
\begin{proof}
Inspired by~\cite{Bu2021,Du2019},
suppose the conclusion fails to hold at time $t$, we can decompose it into three situations (1) $\lambda_{min}(\bm{H}(t)) < \lambda_0/2$, (2) $\|\w_r(t) - \w_r(0)\|_2 > R^{'}$ for all $r \in [m]$ or (3) $\ell(t) > \frac{6\hat{\ell}(0)}{\lambda_0} e^{-(2-\sqrt{2})\sqrt{\frac{\lambda_0}{2}} t}$.

When $\lambda_{min}(\bm{H}(t)) < \lambda_0/2$, based on Lemma~\ref{lem:B2}, it has
$\|\w_r(t) - \w_r(0)\|_2 > R$.
Thus, 
there exists a $t_0$ that 
\begin{equation}
	t_0 = \inf\{t^{'}:\max_{r \in [m]}\|\w_r(t^{'}) - \w_r(0)\|_2 \geq R\}.
\end{equation} 
As a result, it exists a $r \in [m]$ satisfied $\|\w_r(t_0) - \w_r(0)\|_2=R$.
Applying Lemma~\ref{lemma3.3}, it has $\lambda_{min}(H(t^{'})) > \lambda_0/2$ for $t^{'} \leq t_0$.
According to Lemma~\ref{lemma:4.1}, it has $\|\w_r(t_0) - \w_r(0)\| \leq R^{'} < R$, which makes a contradiction.

Similarly, it exists a $i$ that $\lambda_{min}(\bm{H}(i)) < \lambda_0/2$ for case (2) or (3) according to Lemma~\ref{lemma3.3}.
The rest of the proof is similar as case (1).
\end{proof} 

\subsection{Proof of Lemma~\ref{lemma:4.3}}
\label{proof4.3}
\begin{proof}
Motivated by~\cite{Shi2021}, we consider the Lyapunov function
\begin{eqnarray}
\label{lemma6_V}
	V(t) &=& (1+\sqrt{\frac{\lambda_0 s}{2}})\hat{\ell}(t) + \frac{1}{4}\|\dot{\Delta}(t)\|^2 + \frac{1}{4}\|\dot{\Delta}(t)+\sqrt{2\lambda_0}\Delta(t) + \sqrt{s}\bm{H}(t) \Delta(t)\|^2 \nonumber \\
&\leq&(1+\sqrt{\frac{\lambda_0 s}{2}})\hat{\ell}(t) + \frac{1}{4}\|\dot{\Delta}(t)\|^2 +  \frac{1}{4}(1+\frac{1}{\phi})\|\dot{\Delta}(t)\|^2 + \frac{1}{4}(1+\phi)\|\sqrt{2\lambda_0}\Delta(t) + \sqrt{s}\bm{H}(t)\Delta(t)\|_2^2
\nonumber\\
		&\leq& (1+\sqrt{\frac{\lambda_0 s}{2}})(3+2\phi)\hat{\ell}(t) + \frac{1}{4}(2+\frac{1}{\phi})\|\dot{\Delta}(t)\|^2   +\frac{s(1+\phi)}{4}\|\bm{H}(t)\Delta(t)\|^2,  
\end{eqnarray}
where the first inequality uses Young's inequality and the second inequality exploits $\frac{\lambda_0}{2}\|\Delta(t)\|^2\leq 2\hat{\ell}(t)$.

The corresponding derivative of V has the bound
\begin{eqnarray}
	\dot{V}(t) 
	&=& (1+\sqrt{\frac{\lambda_0 s}{2}})\big(\dot{\Delta}(t)^{\top}\bm{H}(t)\Delta(t) + \frac{1}{2} \Delta(t)^{\top}\dot{\bm{H}}(t)\Delta(t) \big) + \frac{1}{2}\langle \dot{\Delta}(t),\ddot{\Delta}(t)\rangle \nonumber \\
	&&+ \frac{1}{2}\langle \dot{\Delta}(t)+\sqrt{2\lambda_0}\Delta(t) + \sqrt{s}\bm{H}(t) \Delta(t), \ddot{\Delta}(t)+\sqrt{2\lambda_0}\dot{\Delta}(t) + \sqrt{s}\dot{\bm{H}}(t) \Delta(t)+\sqrt{s}\bm{H}(t)\dot{\Delta}(t)\rangle \nonumber \\
	&=& -\frac{\sqrt{2\lambda_0}}{2}\|\dot{\Delta}(t)\|^2 -\frac{\sqrt{s}}{2} \dot{\Delta}^{\top}(t)\bm{H}(t)\dot{\Delta}(t)-(1+\sqrt{\frac{\lambda_0 s}{2}})\sqrt{2\lambda_0}\hat{\ell}(t) - \frac{\sqrt{s}}{2}(1+\sqrt{\frac{\lambda_0 s}{2}})\|\bm{H}(t)\Delta(t)\|^2 \nonumber\\
\label{C.1}	\\
	&\leq&  -(1+\sqrt{\frac{\lambda_0 s}{2}})\sqrt{2\lambda_0}\hat{\ell}(t) -\frac{\sqrt{2\lambda_0}}{2}(1+\frac{\sqrt{2\lambda_0 s}}{4})\|\dot{\Delta}(t)\|^2  - \frac{\sqrt{s}}{2}(1+\sqrt{\frac{\lambda_0 s}{2}})\|\bm{H}(t)\Delta(t)\|^2 \nonumber\\
&\leq&  -(1+\sqrt{\frac{\lambda_0 s}{2}})\sqrt{2\lambda_0}\hat{\ell}(t) -\frac{\sqrt{2\lambda_0}}{2}(1+\frac{\sqrt{2\lambda_0 s}}{4})\|\dot{\Delta}(t)\|^2  - s\sqrt{\frac{\lambda_0 }{2}}\|\bm{H}(t)\Delta(t)\|^2, \nonumber
\end{eqnarray}
where the first inequality uses the positive definite of $\bm{H}(t)$ and the last inequality uses $1\geq \sqrt{\frac{\lambda_0 s}{2}}$.
Thus, with the bounds of $V(t)$ and $\dot{V}(t)$, NAG has 
\begin{eqnarray}
\label{convergence_rate_new}
\rho_{NAG}^* = \max_{\phi>0} \min\{ \frac{2}{3+2\phi}, \frac{4(1+\frac{\sqrt{2\lambda_0 s}}{4})}{2+1/\phi}, \frac{4}{1+\phi}\} \sqrt{\frac{\lambda_0}{2}}. \nonumber 
\end{eqnarray}
Let $\alpha = \frac{\sqrt{2\lambda_0 s}}{4}$, which has a bound $0\leq \alpha \leq \frac{1}{2\sqrt{\kappa}}\leq 1/2$.
Then it has
\begin{eqnarray}
	\rho_{NAG}^*(\alpha) = \frac{1}{2}(4+3\alpha-\sqrt{8+16\alpha+9\alpha^2}),
\end{eqnarray}
which is a monotonous increasing function with respect to $\alpha$ when $\alpha > 0$.

Therefore, it has
\begin{eqnarray}
\label{NAG: ode_result}
V(t) \leq e^{-\rho_{NAG}^*(\alpha)\sqrt{\lambda_0/2} t} V(0).
\end{eqnarray}

With the initial value $\dot{w}_r(0)= 0$ for any $r \in [m]$, it has
\begin{eqnarray}
	V(t) \leq e^{-\rho_{NAG}^*(\alpha)\sqrt{\lambda_0/2} t} V(0). \nonumber
\end{eqnarray}
As a result, it has
\begin{eqnarray}
	(1+\sqrt{\frac{\lambda_0 s}{2}})\hat{\ell}(t) &\leq& V(t) \nonumber\\
	&\leq& e^{-\rho_{NAG}^*(\alpha)\sqrt{\lambda_0/2} t}\big( (1+\sqrt{\frac{\lambda_0 s}{2}})\hat{\ell}(0) + \frac{1}{4}\|\sqrt{2\lambda_0}\Delta(0)+\sqrt{s}\bm{H}(0)\Delta(0)\|^2 \big) \nonumber \\
	&\leq& e^{-\rho_{NAG}^*(\alpha)\sqrt{\lambda_0/2} t}\big( (1+\sqrt{\frac{\lambda_0 s}{2}})\hat{\ell}(0) + \frac{\lambda_0}{2} \|\Delta(0)\|_2^2 + \frac{s}{4}\|\bm{H}(0)\Delta(0)\|_2^2 + \sqrt{2\lambda_0 s}\hat{\ell}(0)\big) \nonumber \\
	&\leq& e^{-\rho_{NAG}^*(\alpha)\sqrt{\lambda_0/2} t}\big( (1+\sqrt{\frac{\lambda_0 s}{2}})\hat{\ell}(0) + \frac{4}{3}\hat{\ell}(0) + 2\hat{\ell}(0)\big), \nonumber
\end{eqnarray}
where the last inequality uses $s\leq 1/\lambda_m$, $\|\bm{H}(0)\Delta(0)\|_2^2 \leq \Delta(0)^{\top}\bm{H}(0)\Delta(0)\|\bm{H}(0)\|_2\leq 2\hat{\ell}(0)\lambda_m$ and  $\hat{\ell}(0) \geq \frac{3\lambda_0}{4}\|\Delta(0)\|_2^2$.
Then, it has
\begin{eqnarray}
	\hat{\ell}(t) &\leq& e^{-\rho_{NAG}^*(\alpha)\sqrt{\lambda_0/2} t}  \frac{13/3+\sqrt{\frac{\lambda_0 s}{2}}}{1+\sqrt{\frac{\lambda_0 s}{2}}} \hat{\ell}(0) \nonumber \\
	\ell(t) &\leq& \frac{26\hat{\ell}(0)}{3\lambda_0} e^{-\rho_{NAG}^*(\alpha)\sqrt{\lambda_0/2} t}.
\end{eqnarray}

Now, we turn to prove the bound of the distance between  $\w_r(t)$ and $\w_r(0)$.
Based on Eq.(\ref{eq:NAG}), it has
\begin{eqnarray}
	&&\frac{d}{dt}\left(e^{\sqrt{2\lambda_0}t}\big(\dot{\w}_r + \sqrt{s}( \frac{\partial \f}{\partial \w_r})^{\top}(\f-\y)\big)\right) \nonumber \\
	&=& \sqrt{2\lambda_0}e^{\sqrt{2\lambda_0}t}\big(\dot{\w}_r + \sqrt{s}( \frac{\partial \f}{\partial \w_r})^{\top}(\f-\y)\big) + e^{\sqrt{2\lambda_0}t}\big( \ddot{\w}_r+ \sqrt{s}( \frac{\partial \f}{\partial \w_r})^{\top}\dot{\f} \big)
	\nonumber \\
	&=& e^{\sqrt{2\lambda_0}t}\big(-(1+\sqrt{\frac{\lambda_0 s}{2}})\frac{\partial \ell(\W, \bm{a})}{\partial \w_r} + \sqrt{2\lambda_0 s}( \frac{\partial \f}{\partial \w_r})^{\top}(\f-\y) \big),
	\nonumber 
\end{eqnarray}
where the first equality uses Eq.(\ref{eq:16}) and Eq.(\ref{eq:18}).

By integrating both sides of the above formula, it has 
\begin{eqnarray}
\label{distance_HB}
	&&\dot{\w}_r(t) \nonumber\\
	&=&\!\!\!\!\!-\sqrt{s}( \frac{\partial \f(t)}{\partial \w_r(t)})^{\top}(\f(t)-\y) \!+\!e^{-\sqrt{2\lambda_0}t} \int_0^t e^{\sqrt{2\lambda_0}t^{'}}(-(1\!+\!\sqrt{\frac{\lambda_0 s}{2}})\frac{\partial \ell(\W(t^{'}), \bm{a})}{\partial \w_r(t^{'})} + \sqrt{2\lambda_0 s}( \frac{\partial \f(t^{'})}{\partial \w_r(t^{'})})^{\top}(\f(t^{'})-\y))dt^{'}. \nonumber
\end{eqnarray}

Taking the norm and applying $\sum_{i=1}^n \|z_i\|_2 \leq \sqrt{n} \|\bm{z}\|_2$, then it has
\begin{eqnarray}
	\|\dot{\w}_r(t)\| &\leq& \sqrt{\frac{sn}{m}}\|\f(t)-\y\|_2 +  e^{-\sqrt{2\lambda_0}t} \int_0^t e^{\sqrt{2\lambda_0}t^{'}}(1+\frac{3\sqrt{2\lambda_0 s}}{2})\frac{\sqrt{n}}{\sqrt{m}}
\|\f(t^{'}) - \y\|_2 dt^{'} \nonumber\\
	&\leq& \sqrt{\frac{52sn\hat{\ell}(0)}{3\lambda_0 m}}e^{- \frac{\rho_{NAG}^*(\alpha)}{2}\sqrt{\frac{\lambda_0}{2}}t} 
	+(1+\frac{3\sqrt{2\lambda_0 s}}{2})\sqrt{\frac{52n\hat{\ell}(0)}{3\lambda_0 m}} e^{-\sqrt{2\lambda_0}t}\frac{e^{(\sqrt{2\lambda_0} - \frac{\rho_{NAG}^*(\alpha)}{2}\sqrt{\frac{\lambda_0}{2}})t} - 1}{\sqrt{2\lambda_0} - \frac{\rho_{NAG}^*(\alpha)}{2}\sqrt{\frac{\lambda_0}{2}}} \nonumber \\
	&\leq&  3.33\sqrt{\frac{52\hat{\ell}(0) n}{3\lambda^2_0 m}}e^{-\frac{\sqrt{2}-1}{2}	\sqrt{\lambda_0}t},\nonumber
\end{eqnarray}
where the first inequality applies $\|\frac{\partial f_i(t)}{\partial \w_r(t)}\|_2 \leq 1/\sqrt{m}$ for any $i \in [n]$ and the last inequality uses $s\leq 1/\lambda_m$, $\lambda_m = \lambda_{max}(\bm{H}^{\infty}) + \frac{\lambda_0}{4} \geq \frac{5\lambda_0}{4}$.
Applying Cauchy-Schwarz inequality, it has
\begin{eqnarray}
	\|\w_r(t) - \w_r(0)\|_2 \leq \int_0^t \|\dot{\w}_r(t^{'})\|_2 d t^{'} \leq 67\sqrt{\frac{\hat{\ell}(0) n}{\lambda^3_0 m}}. \nonumber
\end{eqnarray}
\end{proof}
\begin{figure*}[t]
\centering
\includegraphics[width=0.99\textwidth]{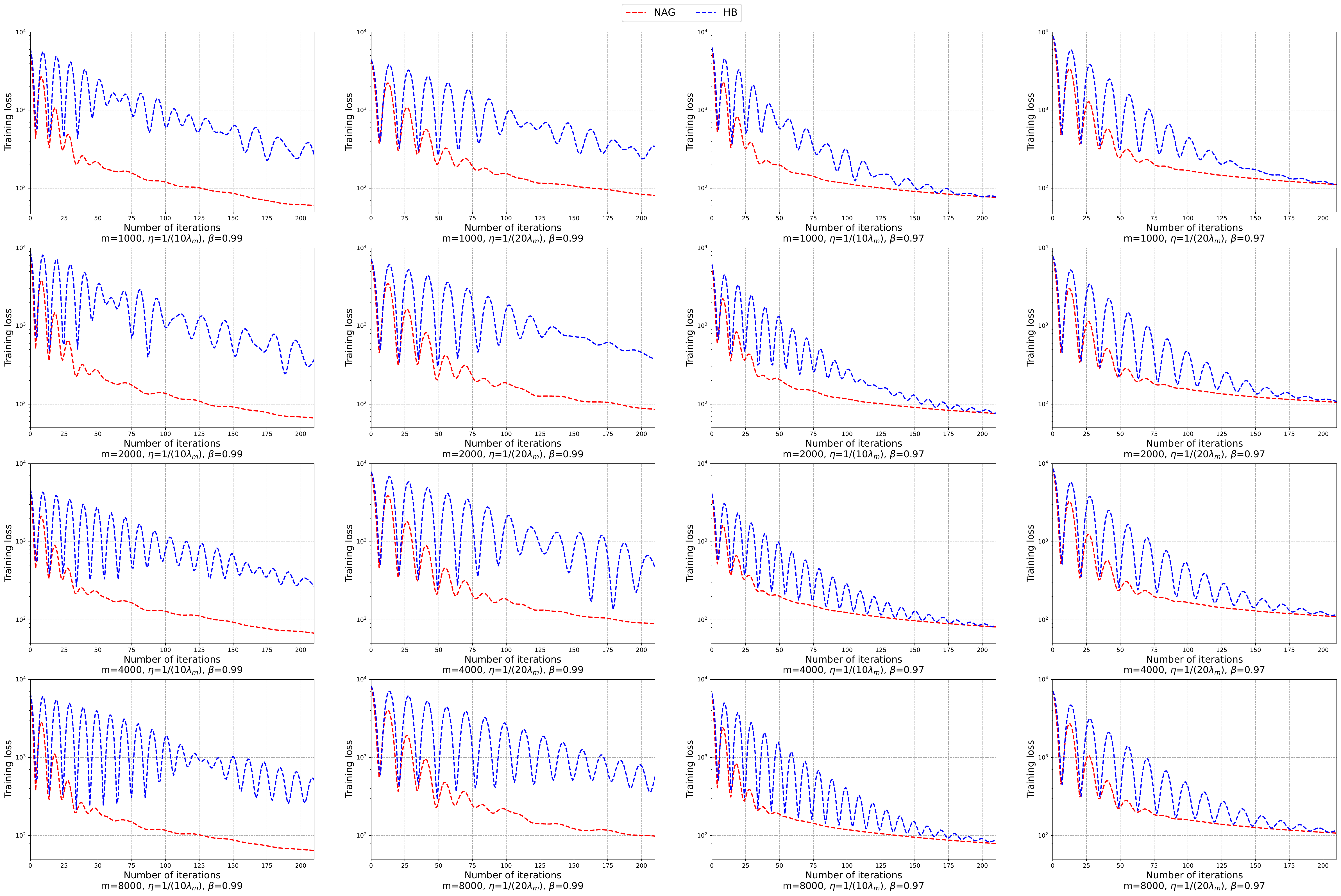} 
\caption{Convergence comparison between NAG and HB in training two-layer ReLU neural networks on the FMNIST dataset with different width $m$, learning rate $\eta$ and momentum parameter $\beta$.  }
\label{fig2.1}
\end{figure*}
\subsection{Additional experiments}
\label{additional experiments}
\subsubsection{Convergence comparison.}
In this subsection, we provide additional convergence comparison results on FMNIST and CIFAR10 datasets.
As observed in Fig.~\ref{fig2.1} and Fig.~\ref{fig2.2}, NAG converges faster than HB in training two-layer ReLU neural networks.

\begin{figure*}[!h]
\centering
\includegraphics[width=0.99\textwidth]{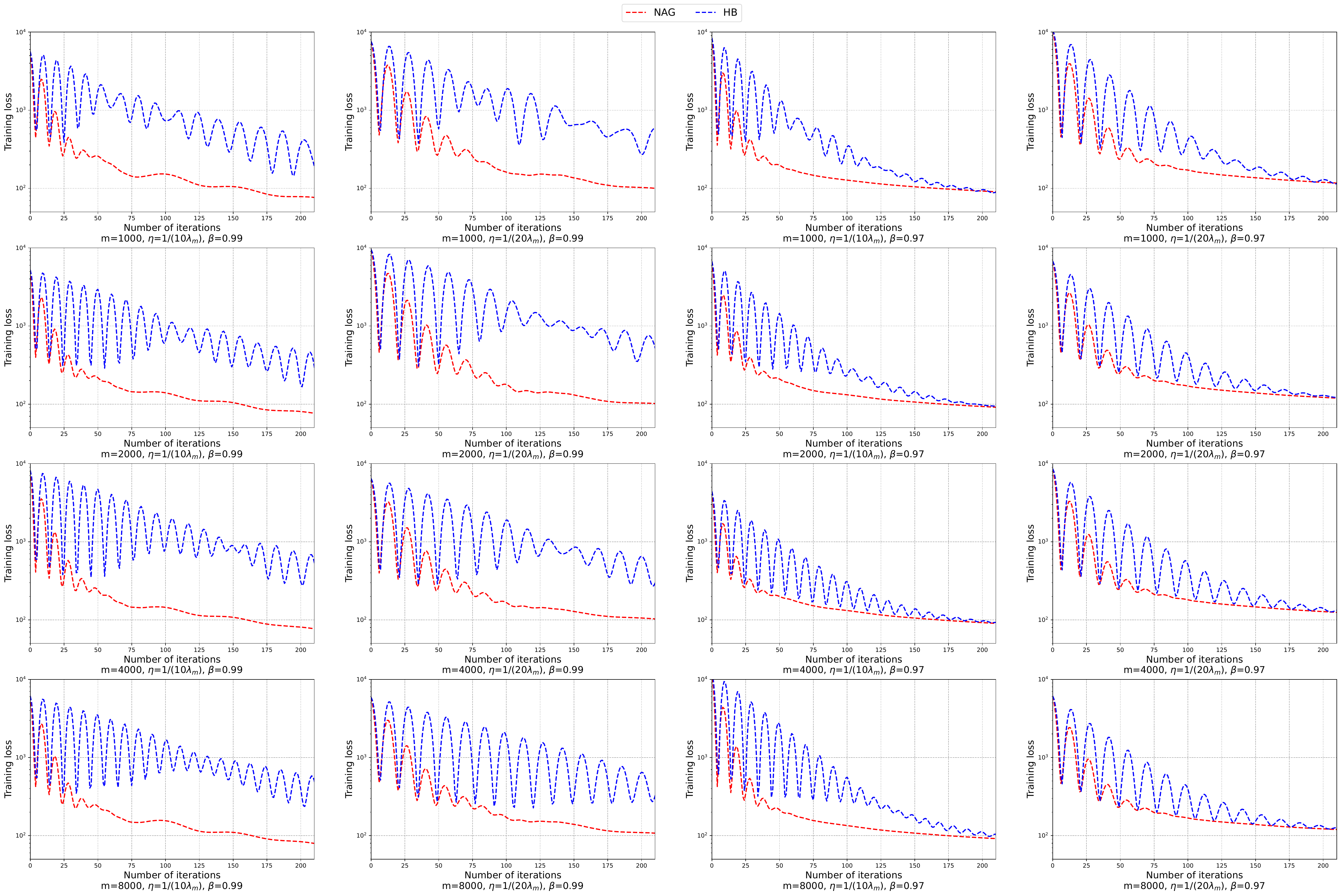} 
\caption{Convergence comparison between NAG and HB in training two-layer ReLU neural networks on the CIFAR10 dataset with different width $m$, learning rate $\eta$ and momentum parameter $\beta$.}
\label{fig2.2}
\end{figure*}

\subsubsection{Maximum distance.}
We also provide additional experiments on the maximum distance $\max_{r \in [m]}\|\w_r(t) - \w_0(t)\|$ for HB and NAG on FMNIST and CIFAR10 datasets, where the results are provided in Fig.~\ref{fig3.1}-\ref{fig4.2}.
\begin{figure*}[!h]
\centering
\includegraphics[width=0.99\textwidth]{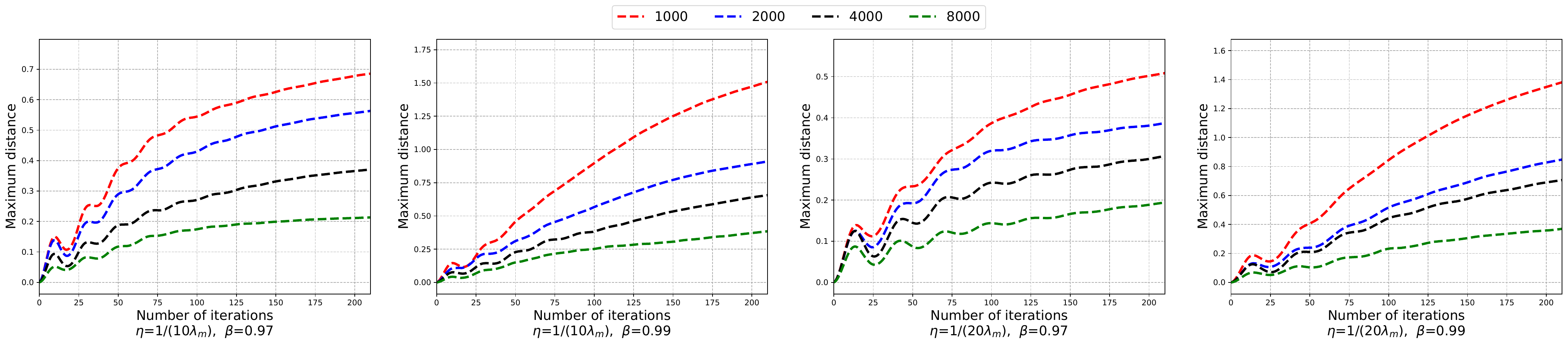} 
\caption{The maximum distance $\max_{r \in [m]}\|\w_r(t) - \w_0(t)\|$ of HB in training two-layer ReLU neural networks on the FMNIST dataset during training process.}
\label{fig3.1}
\end{figure*}

\begin{figure*}[!h]
\centering
\includegraphics[width=0.99\textwidth]{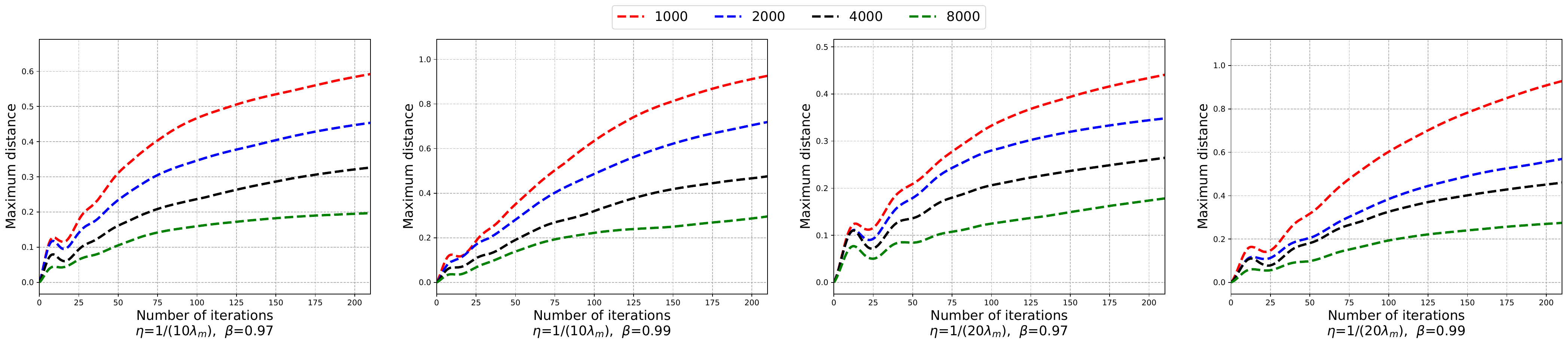} 
\caption{The maximum distance $\max_{r \in [m]}\|\w_r(t) - \w_0(t)\|$ of NAG in training two-layer ReLU neural networks on the FMNIST dataset during training process.}
\label{fig4.1}
\end{figure*}

\begin{figure*}[t]
\centering
\includegraphics[width=0.99\textwidth]{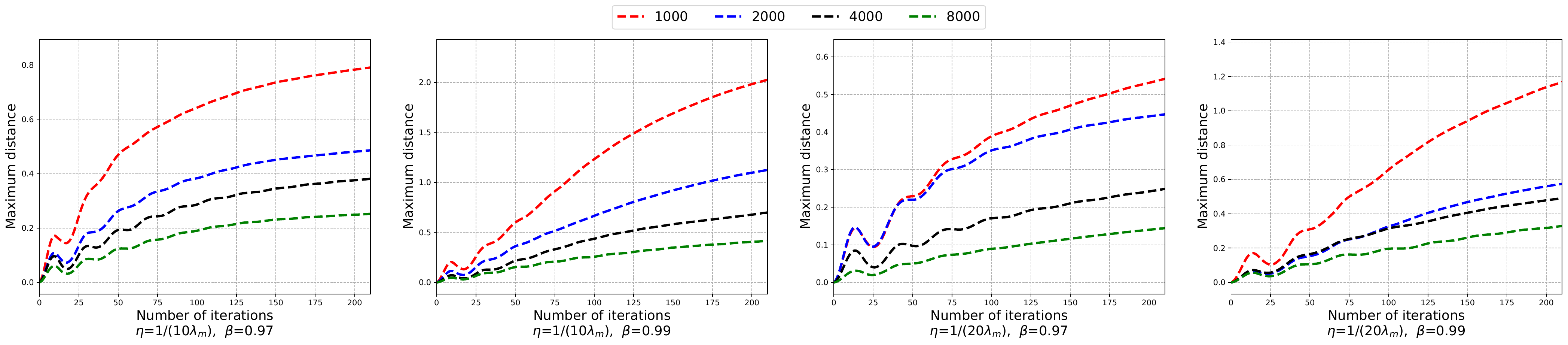} 
\caption{The maximum distance $\max_{r \in [m]}\|\w_r(t) - \w_0(t)\|$ of HB in training two-layer ReLU neural networks on the CIFAR10 dataset during training process.  }
\label{fig3.2}
\end{figure*}

\begin{figure*}[t]
\centering
\includegraphics[width=0.99\textwidth]{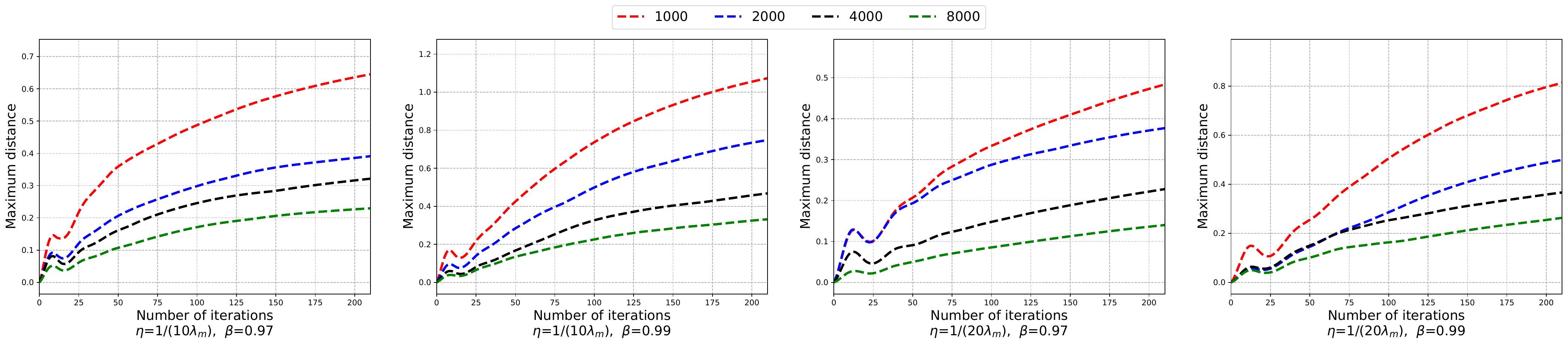} 
\caption{The maximum distance $\max_{r \in [m]}\|\w_r(t) - \w_0(t)\|$ of NAG in training two-layer ReLU neural networks on the CIFAR10 dataset during training process.  }
\label{fig4.2}
\end{figure*}
\end{document}